\documentclass[11pt]{article}
\usepackage{acl}
\usepackage{times}
\usepackage{subfigure}
\usepackage{graphicx}
\usepackage{latexsym}
\usepackage{booktabs} 
\usepackage{multirow}
\usepackage{xcolor}
\usepackage{amsmath,amsfonts,amssymb}

\definecolor{mypurple}{RGB}{148,0,211}
\definecolor{mygreen}{RGB}{34,139,34}
\definecolor{mysteelblue}{RGB}{70,130,180}
\definecolor{myblue}{RGB}{0,0,205}

\usepackage{microtype}

\title{Unified Speech-Text Pre-training for Speech Translation and Recognition}

\author{Yun Tang, Hongyu Gong, Ning Dong, Changhan Wang, \\ \textbf{Wei-Ning Hsu, Jiatao Gu, Alexei Baevski, Xian Li,} \\ \textbf{Abdelrahman Mohamed, Michael Auli, Juan Pino} \\
 Meta AI\\
\texttt{\{yuntang,hygong,dnn,changhan,wnhsu,jgu,abaevski,xianl,}\\\texttt{abdo,michaelauli,juancarabina\}@fb.com}         
}

\date{}

\begin{document}
\maketitle
\begin{abstract}
We describe a method to jointly pre-train 
speech and text in an encoder-decoder modeling framework for speech translation and recognition. 
The proposed method incorporates 
four self-supervised and supervised subtasks for cross modality learning. 
A self-supervised speech subtask leverages unlabelled speech data, and a (self-)supervised text to text subtask makes use of 
abundant text training data.  
Two auxiliary supervised speech tasks are included to unify speech and text modeling space. 
Our contribution lies in integrating linguistic information from the text corpus into the speech pre-training.
Detailed analysis reveals learning interference among subtasks. 
Two pre-training configurations for speech translation and recognition, respectively, are presented to alleviate subtask interference.
Our experiments show the proposed method can effectively fuse speech and text information into one model. 
It achieves between 1.7 and 2.3 BLEU improvement above the state of the art on the \textsc{MuST-C} speech translation dataset and 
comparable WERs to wav2vec 2.0 on the \textsc{Librispeech} speech recognition task.
\footnote{\href{https://github.com/pytorch/fairseq/tree/main/examples/speech_text_joint_to_text}{https://github.com/pytorch/fairseq/tree/main/\\examples/speech\_text\_joint\_to\_text}.}
\end{abstract}
\section{Introduction}


Pre-training can learn universal feature representations from a large training corpus and is beneficial for  downstream tasks with limited amounts of  training data
~\cite{Peters2018DeepCW,Oord2018RepresentationLW,Chung2018UnsupervisedCA,Zoph2020RethinkingPA}.
With the advancement of computational power and self-supervised pre-training approaches, large volumes of unlabeled data may now be used in pre-training.
Methods, such as BERT~\cite{Devlin2019BERTPO}, BART~\cite{lewis-etal-2020-bart} and wav2vec2.0~\cite{Baevski2020wav2vecV2}, have emerged as the backbone of many speech and natural language processing tasks.
 

The aforementioned pre-training methods focus on learning feature representation either from text or speech.  Many speech applications combine information learnt from both speech and text corpora to achieve state of the art results.  In speech processing, transcribed speech training data is generally very scarce for many languages. It is difficult to build robust linguistic knowledge representation solely based on labeled speech training data. 
\citet{Jia2019LeveragingWS,Chen2021InjectingTI} propose to generate synthetic data from text to augment speech training corpus.  \citet{Li2021MultilingualST} demonstrate that models initialized with pre-trained wav2vec2.0 and mBART~\cite{Liu2020MultilingualDP} modules are competitive for the multilingual speech to text translation task.
\citet{Chuang2020SpeechBERTAA} propose to concatenate the acoustic model and BERT model  for speech Q\&A.  \citet{Chung2021SPLATSJ} 
align speech utterance representation to the corresponding text sentence representation,  in which both representations are generated from unsupervised pre-trained models, for speech understanding.

In this study, we are interested in pre-training for speech to text tasks using the Attention based Encoder-Decoder (AED) framework. In particular, we seek to answer the question whether
the integration of data from different modalities is beneficial for representation learning.  
To answer this question, we propose Speech and Text joint Pre-Training (\textsc{STPT}), a multi-task learning framework to combine different modalities, i.e., speech and text, in the pre-training stage. 
A self-supervised speech subtask and a (self-)supervised text to text subtask dominate the pre-training computation to leverage large amounts of unlabelled speech data and abundant text training corpus. 
Two auxiliary supervised speech subtasks are used to unify different modalities in the same modeling space. 
The proposed method fuses information from the text and speech training corpus into a single model, and it effectively improves the performance of downstream tasks, such as speech to text translation (ST) and automatic speech recognition (ASR). 
Our contributions  are summarized as follows:
\begin{enumerate}
    \item We propose a multi-task learning framework to learn four speech and text subtasks in one model and successfully integrate linguistic information from the text corpus into the speech pre-training. 
    \item We conduct detailed analyses on the proposed pre-training method, which reveal
    the interference among different subtasks. 
    \item Two joint pre-training configurations are proposed to alleviate  learning interference for ASR and  ST respectively.
    \item State-of-the-art results are achieved on the downstream tasks.  We obtain at least 1.7 BLEU improvement compared
    with the best \textsc{MuST-C} ST system reported and comparable WERs as wav2vec 2.0 in the \textsc{Librispeech} ASR task.
\end{enumerate}

\section{Related work}
\textbf{Pre-training}:  Self-supervised pre-training is usually optimized with two different criteria:  contrastive loss~\cite{Oord2018RepresentationLW,Chung2020ImprovedSR,Baevski2020wav2vecV2} and masked prediction loss~\cite{Devlin2019BERTPO}. Contrastive loss 
focuses on distinguishing the positive samples from the negative ones given the reference sample and it has achieved great success for speech recognition~\cite{Baevski2020wav2vecV2}. Masked prediction loss has been first studied for natural language processing tasks~\cite{Devlin2019BERTPO,lewis-etal-2020-bart} with subsequent application to speech processing~\cite{Baevski2020vqwav2vecSL,Hsu2021HubertHM}.
\citet{Chung2021W2vBERTCC} combine contrastive loss and masked prediction loss, which shows good performance 
for the downstream ASR task. The optimization of our self-supervised speech task is more related to the masked prediction loss. Instead of predicting the hard discretized label for the masked frames, which is error prone, we use KL divergence to minimize the distribution difference between the same feature frames with and without masking. 
Please refer to \autoref{sec:method:ssl} for more details.

\noindent\textbf{Self-training (or iterative pseudo labelling):} self-training is another widely used approach to take advantage of unlabelled speech data to improve the ASR performance~\cite{kahn2020st,Xu2020IterativePF,Pino2020SelfTrainingFE,zhang2020pushing,Wang2021LargeScaleSA,Xiao2021ContrastiveSL,Wang2021UniSpeechAS}. 
A seed model, which usually is trained with a small amount of supervised speech training data, is employed to generate pseudo labels for the unlabelled speech data. 
The speech data with pseudo labels is augmented into the training dataset to build another model, which is expected to outperform the seed model due to more training data exposure. 
Similar to self-training, we also use small amounts of supervised data to unify the speech and text modeling space. 
However, the self-supervised speech training in this work avoids making hard predictions and uses KL divergence to maximize the mutual information between masked span and observed feature frames. 

\noindent\textbf{Multi-task learning}: Due to data scarcity, multi-task learning is widely adopted to leverage parallel text training data for ST~\cite{Weiss2017SequencetoSequenceMC,Anastasopoulos2018TiedML,Tang2021AGM,Ye2021EndtoendST}. Those methods primarily use supervised speech data sets during multi-task learning, whereas our method can leverage large amounts of unlabeled speech data during the pre-training stage, which has the potential to improve performance even further. 

A concurrent work from \citet{Ao2021SpeechT5UE} 
also proposes to jointly pre-train speech and text for ASR and text to speech application, which is fully unsupervised. 
Our method focuses on taking advantage of the supervised speech data, which could be the same data used for fine-tuning, to improve the joint speech text pre-training. Our results demonstrate the efficacy of supervised speech data in pre-training. 
Another concurrent work is from~\citet{Bapna2021SLAMAU}, which focuses on speech encoder pre-training using both speech and text data. Our method emphasizes the encoder-decoder framework and training both encoder and decoder in the pre-training stage. 

\begin{figure*}
\hfill
\centering
\subfigure[ Fully shared encoder (FSE) for ASR pre-training.]{\includegraphics[width=1.00\columnwidth]{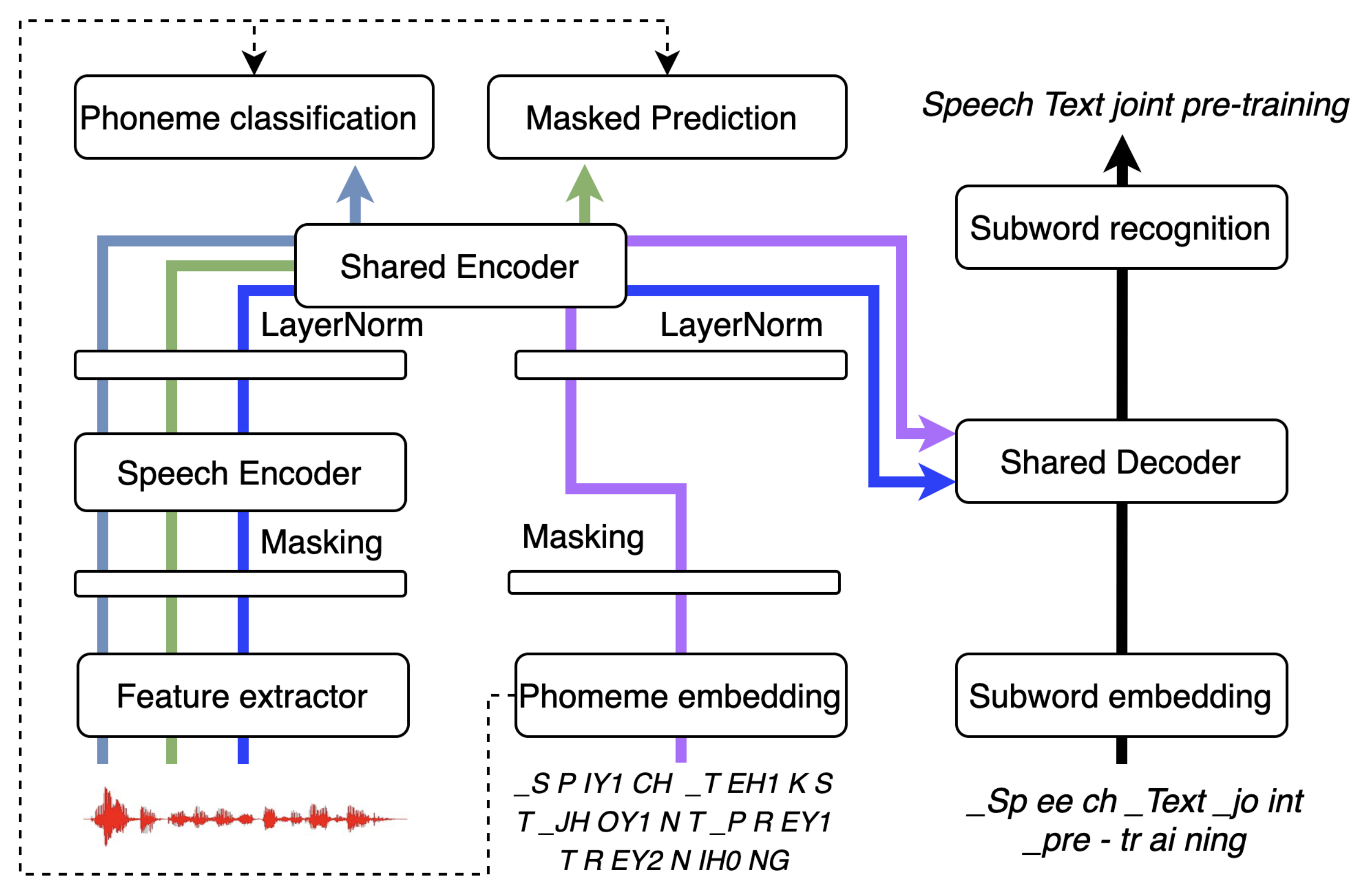}\label{fig:framework_asr}}
\hfill
\subfigure[ Partially shared encoder (PSE) for ST pre-training.]{\includegraphics[width=1.00\columnwidth]{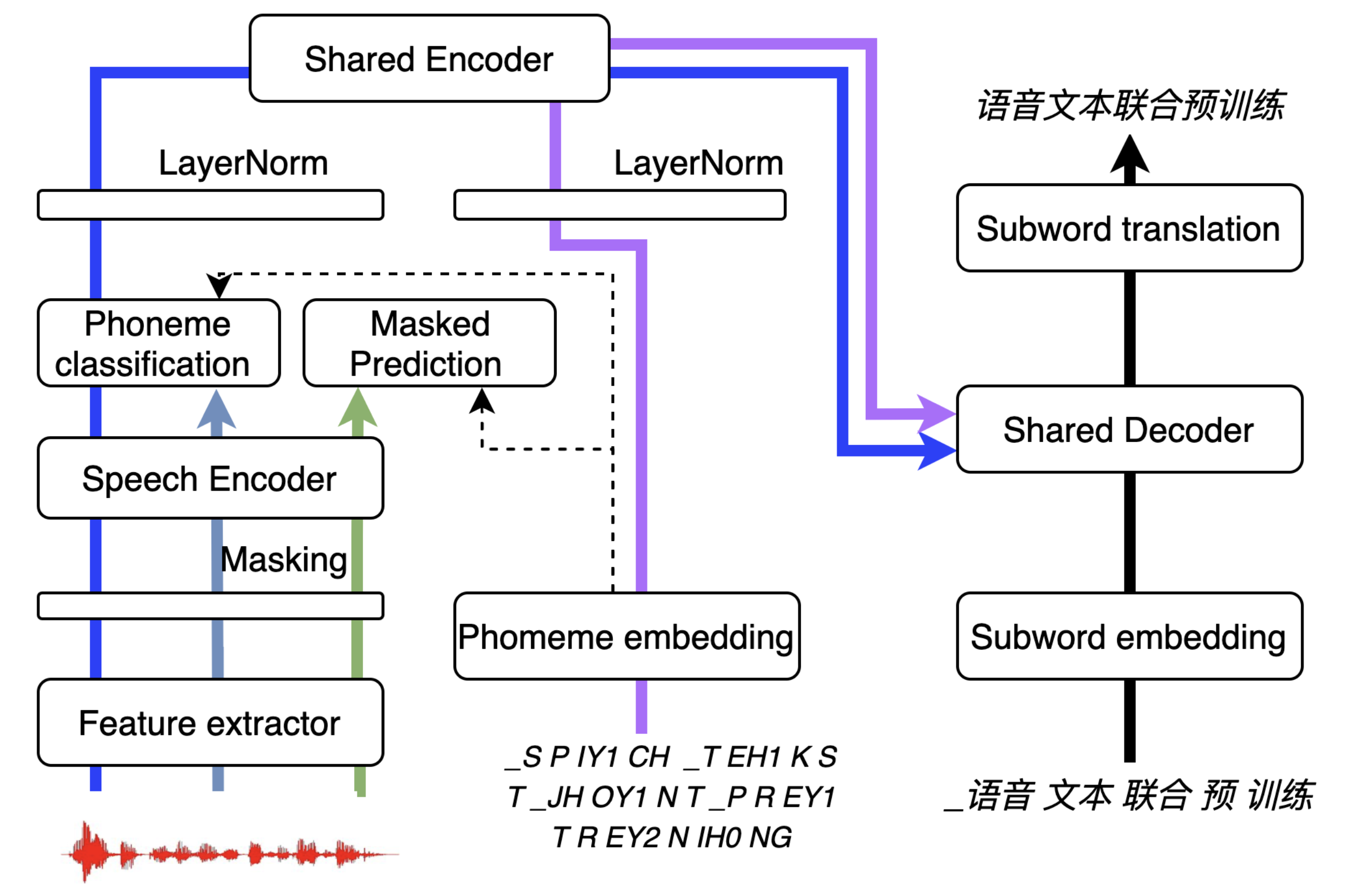}\label{fig:framework_st}}
\hfill
\caption{Speech text joint pre-training framework. The purple, green, steelblue and blue lines depict the data flow in encoders for the \textbf{T}ext to \textbf{T}ext (\textcolor{mypurple}{T2T}), \textbf{S}elf-supervised \textbf{S}peech \textbf{L}earning (\textcolor{mygreen}{SSL}), 
supervised \textbf{S}peech to \textbf{P}honeme classification (\textcolor{mysteelblue}{S2P}) and supervised AED based \textbf{S}peech to \textbf{T}ext (\textcolor{myblue}{S2T}) subtasks respectively. 
The black lines show data flow in the decoder model for the T2T and S2T subtasks. The dotted lines indicate the phoneme embedding is applied in the SSL and S2P subtasks.}
\label{fig:framework}
\end{figure*}

\section{Method}\label{sec:method}
ASR and ST are the two main downstream tasks for the proposed pre-training method. \autoref{fig:framework} depicts our joint pre-training framework, which consists of four subtasks: 
\begin{enumerate}
    \item  (Self-)supervised \textbf{T}ext to \textbf{T}ext subtask (T2T)
    \item \textbf{S}elf-supervised \textbf{S}peech \textbf{L}earning subtask (SSL)
    \item Supervised \textbf{S}peech to \textbf{P}honeme classification subtask (S2P) 
    \item Supervised AED based \textbf{S}peech to \textbf{T}ext subtask, which is the same as the downstream task, i.e., ST or ASR (S2T) 
\end{enumerate}

The choice of the T2T subtask depends on the downstream task.
For ASR, the T2T subtask is a denoising autoencoder task (BART)~\cite{lewis2019bart} while ST utilizes a text based neural machine translation task. 
The SSL subtask is a self-supervised speech learning task to leverage large amounts of unlabelled speech data optimized by the masked prediction loss. 
The last two supervised speech tasks (S2P and S2T) unify two modalities, i.e., speech and text, into one modeling space. 

In this study, we find that the subtasks for the ASR pre-training are complementary, while subtask interference is observed in the ST pre-training at some encoder layers.  
We propose two different configurations: fully shared encoder (FSE) (Figure~\autoref{fig:framework_asr}) for the ASR pre-training, 
and partially shared encoder (PSE) (Figure~\autoref{fig:framework_st}) for the ST pre-training. 
The FSE configuration aims to encourage information sharing between different subtasks while the PSE configuration tries to minimize the information 
sharing between encoder only subtasks, i.e., SSL and S2P, and sequence to sequence AED tasks, i.e., subtask T2T and S2T. 
More subtask interference analysis is presented in \autoref{sec:analysis}.
We describe the details of each subtask in the following subsections. 

\subsection{(Self-)supervised text to text subtask}\label{sec:framework:text_task}
In the sequence to sequence ASR and ST tasks, the decoder is a text generator conditioned on the encoder outputs.  Large amounts of 
training samples are required to cover different linguistic aspects of the target language.  
Abundant text is an ideal supplement to the limited supervised speech data corpus. 
Assume the target text sequence is $Y=(y_1, y_2, \cdots,  y_N)$, its corresponding corrupted version, $X=\textsc{Noise}(Y)=(x_1, x_2, \cdots, x_M)$, can be created by masking or replacing token spans in $Y$~\cite{lewis2019bart} for the ASR pre-training. 
If the downstream task is ST, $X$ is the corresponding source token sequence.  
The task is optimized by maximizing cross entropy 
\begin{equation}
    \mathcal{L}_{\textsc{T2T}} = -\sum_i^N \log p(y_i| y_{1:i-1}, X) 
\end{equation}
In this subtask, we also convert the input text into the corresponding pronunciation form, i.e., phoneme sequence, as it would be easier to align the encoder outputs from speech and text~\cite{Tang2021AGM}.
The purple and black lines in \autoref{fig:framework} describe the data flow in the T2T subtask.

\subsection{Self-supervised speech subtask}\label{sec:method:ssl} 
The SSL subtask aims to leverage vast amounts of unlabelled speech data and learn general speech representations. 
The model configuration follows wav2vec2.0~\cite{Baevski2020wav2vecV2} where the speech model includes a feature extractor and a context encoder. 
The context encoder corresponds to the speech encoder in Figure~\autoref{fig:framework_st} in the ST pre-training. If ASR is the downstream task, the context encoder  
includes one extra shared encoder  as shown in Figure~\autoref{fig:framework_asr}. 
We use different frameworks for the ST and ASR pre-training to reduce interference among subtasks. The detailed subtask interference is discussed in  \autoref{sec:analysis}.

We propose a masked KL divergence loss to optimize the SSL subtask. It consists of two-pass computation. 
Given the speech input $S=(s_1,s_2,\cdots,s_T)$, the feature extractor and context encoder outputs are $Z=(z_1,z_2,\cdots,z_{T'})$ and $O=(o_1,o_2,\cdots, o_{T'})$ respectively, where the speech input is down-sampled by the feature extractor and $T > T'$.  
In the first pass, the output $O$ is compared with the phoneme embedding $E=(e_1, e_2,\cdots,e_I)$, which is from the T2T subtask described in \autoref{sec:framework:text_task}.  $I$ is the phoneme vocabulary size. 
The predicted phoneme distribution $p(o_j | e_i)$ is defined as
\begin{equation}
    p(o_j|e_i) = \frac{\exp({o_j}^\intercal \cdot e_i)}{\sum_{i'} \exp(o_j^\intercal \cdot e_{i'}) }
\end{equation}
In the second pass, speech feature spans $\hat{Z}\subset Z$ are selected and corrupted as wav2vec2.0~\cite{Baevski2020wav2vecV2}. $\hat{O}$ is the corresponding context encoder output from $\hat{Z}$. We train the model to infer the corrupted $p(\hat{o}_j|e_i)$ being similar as $p(o_j|e_i)$  by minimizing KL divergence. 
\begin{equation}
   \mathcal{L}_{\textsc{SSL}} = -\sum_{\hat{o}_j\in \hat{O}} \sum_i p(o_j|e_i)\log \frac{p(\hat{o}_j|e_i)}{p(o_j|e_i)}
\end{equation}\label{equ:unsup_speech}
Compared with the masked prediction loss, instead of predicting the hard discretized label for the masked frames, we use the soft label prediction, i.e., predicted phoneme distribution from the first pass, to learn speech representation and avoid the hard prediction errors. 

\subsection{Supervised speech to phoneme classification}
The S2P subtask is employed to unify the self-supervised trained speech and text models. 
It shares the same model as in the SSL subtask. 
In this subtask, a transcribed ASR data set is used and
the goal of this task is to predict the frame level phoneme labels. 
A HMM-GMM model is trained with the same transcribed dataset using Kaldi~\cite{Povey_ASRU2011} to generate the frame-level labels with forced-alignment.  

The phoneme classification task is optimized with the cross entropy loss, 
\begin{equation}
    \mathcal{L}_{\textsc{S2P}}=- \sum_{o_j \in O} \log p(o_j|e_{a(j)})
\end{equation}
where $a(j)$ is the phoneme label associated with the context encoder output $o_j$.
The data flow in the S2P subtask is depicted with steelblue lines in \autoref{fig:framework}.

\subsection{Supervised AED based speech to text subtask} 
Besides the S2P subtask mentioned in the previous subsection, we include the potential downstream AED based task, i.e. ASR or ST, as another auxiliary subtask during the pre-training stage.  In many speech translation datasets, such as MuST-C~\cite{Gangi2019MuSTCAM} or CoVoST~\cite{Wang2020CoVoST2A}, we have both speech transcription and translation labels. The speech transcription is used in the S2P subtask while the S2T subtask can make use of the corresponding translation labels. 
We hope this auxiliary task would make the transition from pre-training to fine-tuning smooth and result in better performance in downstream tasks.  
The components involved during optimization are connected with blue lines in encoder and black lines in decoder as shown in \autoref{fig:framework}.
They are trained with cross entropy criterion, 
\begin{equation}
    \mathcal{L}_{\textsc{S2T}} = -\sum_t \log p(y_i|y_{i-1}, O)
\end{equation}
where $O$ is the input speech and $Y=(y_1,\cdots,y_N)$ is the target labels. 

The overall pre-training loss is defined as the combination of four losses discussed above
\begin{equation}
    \mathcal{L} = \mathcal{L}_{\textsc{T2T}} + \alpha \mathcal{L}_{\textsc{SSL}} + \beta \mathcal{L}_{\textsc{S2P}} + \gamma \mathcal{L}_{\textsc{S2T}}
\end{equation}
where $\alpha$, $\beta$ and $\gamma$ are task weights for the SSL, S2P and S2T subtasks respectively.

 During the pre-training, the shared encoder inputs come from two sources, 
either from speech encoder outputs in the S2T subtask or phoneme embeddings in the T2T subtask. The shared encoder inputs might be in different numerical scales. In order to stabilize the multi-task training, a LayerNorm~\cite{Ba2016LayerN} is applied
to the shared encoder inputs and places those inputs in the same numerical scale as shown in \autoref{fig:framework}.


\section{Experimental setting}
In the pre-training, we first train modules with the T2T subtask until they are converged. It helps to stabilize the training and achieve a better result.  
Then the entire model is jointly optimized with all subtasks mentioned in \autoref{sec:method}. Finally, the pre-trained model is fine-tuned on the downstream tasks.
In the fine-tuning stage, we keep optimizing the model with the T2T and S2T subtasks.  Two encoder-only subtasks (SSL and S2P) are dropped, since the model has learnt good speech representation from the unlabeled speech data in  pre-training. 

Two downstream tasks, ASR and ST, are examined. 
The ASR system is evaluated on four \textsc{Librispeech}~\cite{Panayotov2015LibrispeechAA} evaluation sets: dev-clean, dev-other, test-clean and test-other.
WER is reported in the experiments.
ST models are evaluated on two translation directions: English-Spanish (EN-ES) and English-French (EN-FR).  Case-sensitive detokenized \textsc{sacrebleu}~\cite{post-2018-call} is reported on the tst-COMMON testset from \textsc{MuST-C}~\cite{Gangi2019MuSTCAM}.    
 
For both ASR and ST pre-training, 60k hours of unlabelled English speech data from Libri-light~\cite{librilight} is used to build the self-supervised speech task if not specifically mentioned.
We employ the same labelled data for the supervised learning in pre-training and fine-tuning, i.e., \textsc{Librispeech} training data for ASR and \textsc{MuST-C} for ST. 
For ASR pre-training, the \textsc{Librispeech} language model~(LM) training dataset is used to build the monolingual BART model. For ST pre-training, we take the parallel training corpus from WMT. More details about the training data could be found in \autoref{sec:data}.
 
\subsection{Model configuration}
The model takes raw speech audio as input. The feature encoder contains seven blocks and the temporal convolutions in each block have 512 channels with strides (5,2,2,2,2,2,2) and kernel widths (10,3,3,3,3,2,2). The speech encoder, shared encoder and shared decoder are all with 6 transformer layers, model dimension 768, inner dimension (FFN) 3,072 and 8 attention heads. We adopt Pre-LN in the transformer block as~\citet{Xiong2020OnLN}.  The total number of parameters is 169 million.

The task weight for each subtask is set by the number of mini-batches used during training.  In the pre-training, the ratio of mini-batch numbers for each subtasks are 1.0,  7.0,  0.5 and 0.5 for the T2T, SSL, S2P and S2T subtasks respectively.   

We mask 30\% tokens in the T2T BART subtask in ASR pre-training, and no masking is applied for the T2T NMT subtask in the ST pre-training. 7\% of the feature frames in the SSL subtask and 3\% of the feature frames in the two supervised speech subtasks are randomly selected as the mask span starting  time-step. The mask span length is 10.  The masking percentage is selected via grid search ($(20,30)$ for text masking, $(6,6.5,7)$ and $(2,3)$ for speech masking).
Additional experimental details such as optimization hyper-parameters are included in \autoref{sec:app:opt}. 

\begin{table*} 
    \centering
    \small
    \begin{tabular}{l| c| c|c|c|c| c}
    \toprule
    \multirow{2}{*}{Data set} & Unlabeled &\multicolumn{2}{c|}{Dev} & \multicolumn{2}{c|}{Test} &  \multirow{2}{*}{ave.}  \\
    \cline{3-6}
      & data & clean & other & clean & other  &    \\
    \hline
     wav2vec 2.0~\cite{Baevski2020wav2vecV2} & LS-960  & 3.2  \; (1.8) & 8.9 \; (4.7) & 3.4 \; (2.1) & 8.5 \; (4.8) & 6.0 \; (3.4) \\
    \hline
    \hline 
     LAS~\cite{park2019specaugment} & - &  - & - &2.8 \; (2.5) &6.8 \; (5.8) & - \\ 
     Transformer~\cite{Tang2021AGM} & - & 2.8 & 7.0 & 3.1 & 7.2 & 5.0  \\ 
    \hline
    \hline 
      STPT & LS-960 &   2.1 \; (1.9) & 5.4  \; (5.2) & 2.3 \; (2.2) &  5.6 \; (5.3) & 3.8 \; (3.6) \\ 
      STPT & LV-60k &  2.0 \; (2.1) &  4.4 \; (4.2) & 2.1 \; (2.1) & 4.6 \; (4.5) & 3.3 \; (3.2) \\ 
    \bottomrule
    \end{tabular}
    \caption{ WER results on Librispeech. ``()'' indicates the WER is measured with an external LM. }\label{tab:asr}
\end{table*}

\begin{table}
    \centering
    \small
    \begin{tabular}{l|c|c}
    \toprule
    {Data corpus} & {EN-ES} &{EN-FR} \\
    \hline
    \citet{Inaguma2020ESPnetSTAS} &  28.0 & 32.7 \\
    \citet{Tang2021ImprovingST} &    31.0  &   37.4  \\
    \citet{Zheng2021FusedAA} &30.8  & - \\ 
    \citet{Ye2021EndtoendST} & 30.8 & 38.0 \\
    \hline\hline
     STPT   &   33.1   &   39.7    \\ 
    \bottomrule
    \end{tabular}
    \caption{ BLEU results of two language pairs on the MuST-C tst-COMMON. }\label{tab:ast}
\end{table}

\section{Experimental results}\label{sec:expt}
\subsection{Main results}
We present the \textsc{Librispeech} recognition results in \autoref{tab:asr}. 
Recognition results without/with an decoding LM are reported. The WERs obtained with LM are displayed within ``()''. 
The second column shows the dataset used as unlabeled data in pre-training. ``LS-960'' stands for \textsc{Librispeech} training dataset and ``LV-60k'' is the 60,000 hours Librilight dataset.
The decoding LM is built with the \textsc{Librispeech} text training corpus , which is the text corpus used by the T2T subtask in the ASR pre-training and fine-tuning.  

The first part of the table shows results from the wav2vec 2.0 base model, which is a CTC based ASR system. Second part of the table presents results from two AED based ASR systems, 
and we mainly compare the proposed method with those two AED systems.
LAS is a LSTM based system trained with the \textsc{Librispeech} data only. 
Transformer~\cite{Tang2021AGM} is based on multi-task learning and jointly trained with a text task.  

The results from STPT models are presented in the third part of the table.  The fourth row shows results from a model that uses 960 hours 
\textsc{Librispeech} training data as the unlabelled pre-training data while the model in the fifth row is pre-trained with the 60k  hours Librilight data. 
STPT outperforms all previous reported AED-based systems. 
On average, there is a 1.2 absolute WER reduction obtained compared to the jointly trained transformer model~\cite{Tang2021AGM}. STPT also reduces 2.2 WER compared with the CTC based wav2vec model if no external LM is applied and achieves comparable WERs  when it is decoded with a LM.
One interesting observation is the decoding LM is not very helpful for the STPT model, that 
 only 0.2 WER reduction is observed when a decoding LM is applied. 
Other systems, on the other hand, show a considerable WER reduction when the LM is applied during decoding.
It indicates that our multi-task learning in the pre-training and fine-tuning stages can effectively fuse linguistic information in the text data corpus into the ASR model.   LM might not be required if it is trained on the same text corpus.  
We also report results from model pre-trained with 60k hours Librilight data at the fifth row. Compared with the LS-960 STPT model, Librilight data helps to reduce the WER in two difficult ``other'' datasets. In the following experiments, we will use Librilight as unlabelled data in pre-training.

In \autoref{tab:ast}, we present the speech translation results on the MuST-C datasets. Row one to four are the latest results from literature.  Row one shows the results by training a speech to text translation task alone.
Row two and three present results from two multi-task systems with speech and text jointly trained together. Row four is the best system reported, which is initialized with the pre-trained wav2vec 2.0 and machine translation model,  then fine-tuned with the joint speech and text training.
Our method achieves  {2.3} and {1.7} more BLEU scores for EN-ES and EN-FR translation directions compared with the best system~\cite{Ye2021EndtoendST}.

\begin{figure*}[t]
\hfill
\centering
\subfigure[Gradient similarity for the ASR pre-training.]{\includegraphics[width=1.0\columnwidth]{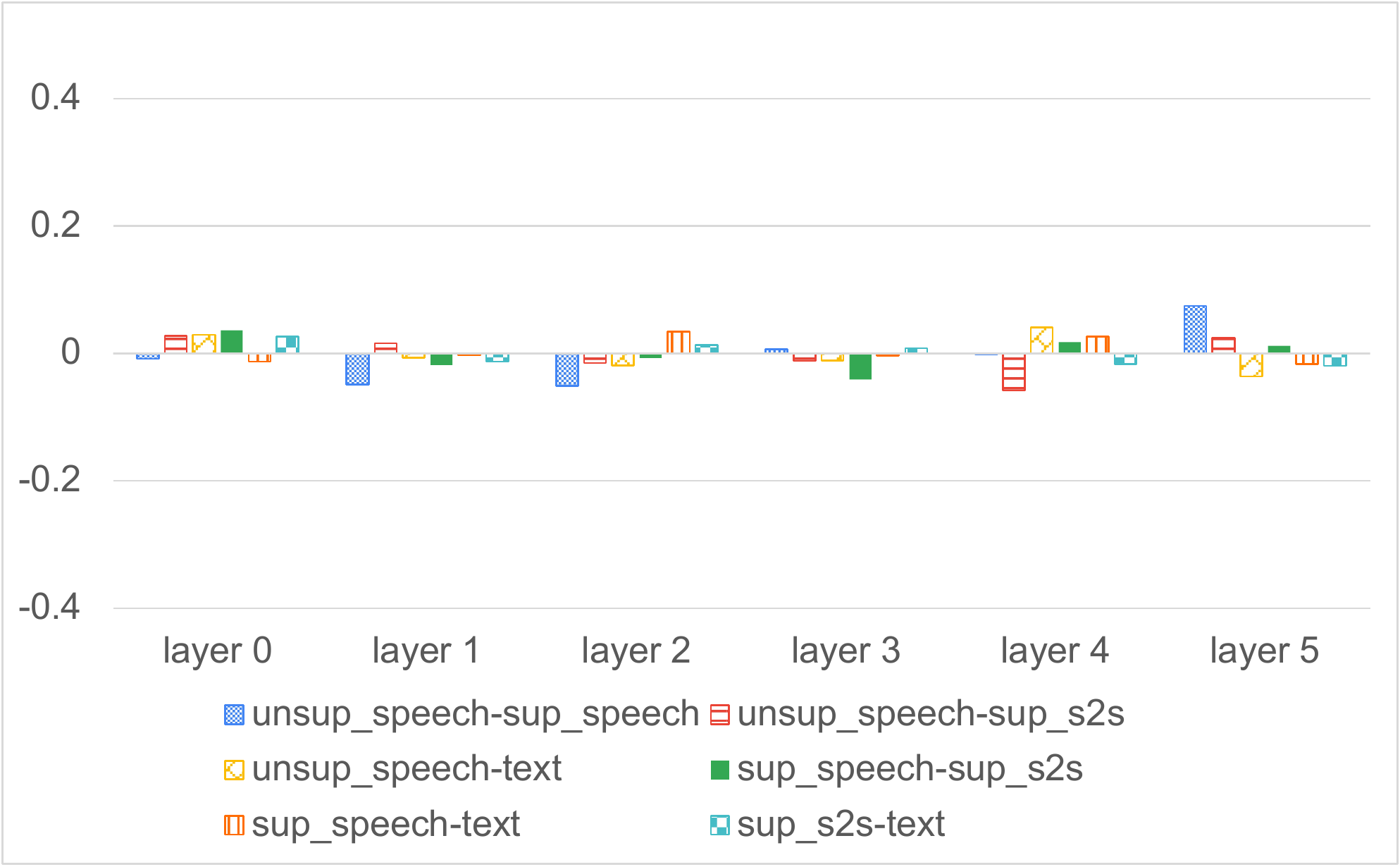}\label{fig:tc:asr}}
\hfill
\subfigure[Gradient similarity for the ST pre-training.]{\includegraphics[width=1.0\columnwidth]{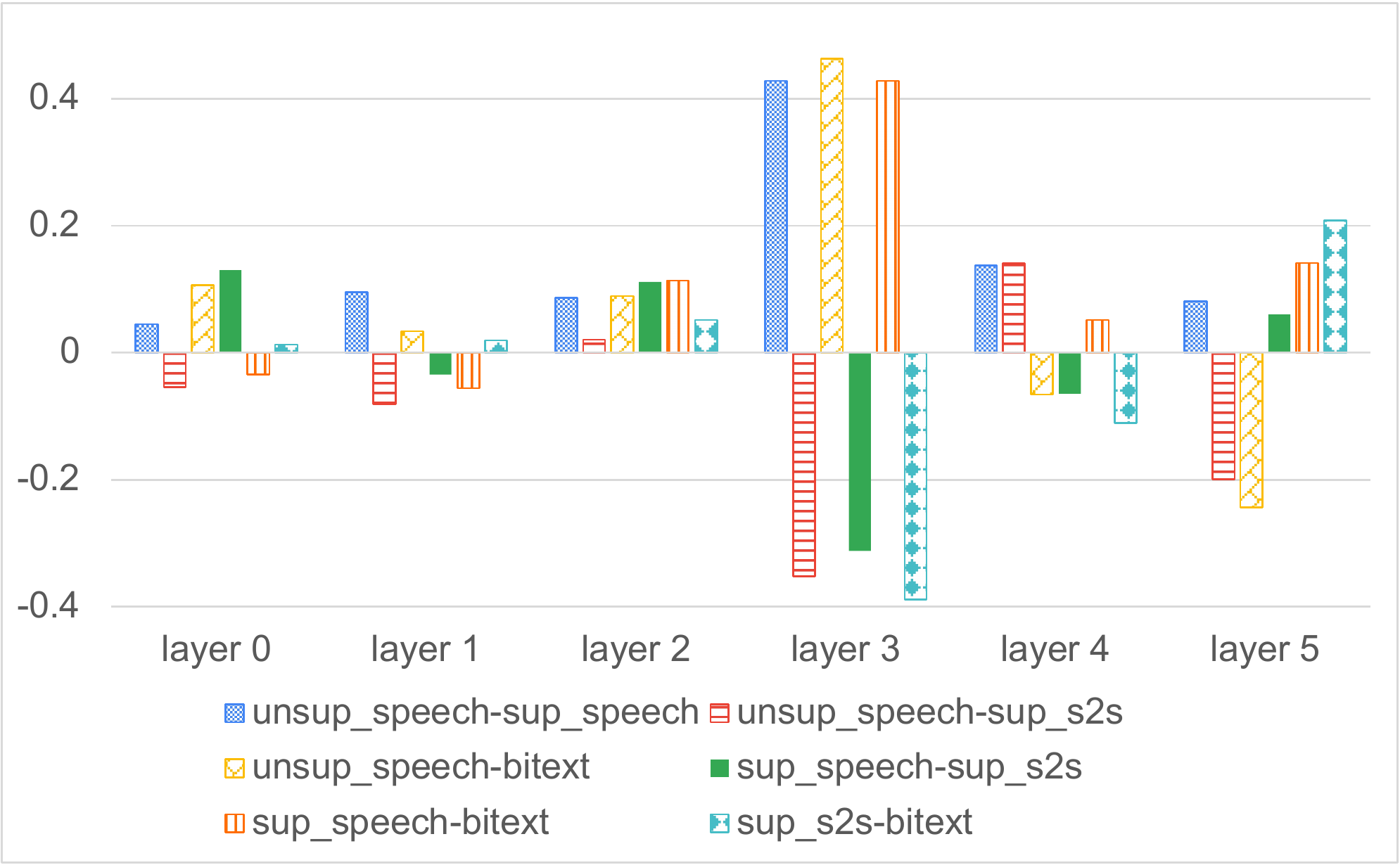}\label{fig:tc:st}}
\hfill
\caption{Gradient similarity for different subtasks on the shared text encoder.  
}
\label{fig:task_conflicts}
\end{figure*}

\subsection{Impact of model structure}\label{sec:analysis}
Interference among subtasks may impede the progress of multi-task learning and  lead to inferior results. In this study, 
we examine the task interference via comparing the gradient similarity between pair subtasks. We choose the pre-trained models using the FSE configuration  discussed in~\autoref{sec:method} and accumulate gradients from one of four jointly trained subtasks. 
We prepare $20$ batches of training samples for each subtask, and retrieve the accumulated gradients by sending these batches to the models.
Then we calculate the pairwise cosine similarity between gradients from any two subtasks. 

The pairwise subtask gradient similarity from the shared encoder are presented in~\autoref{fig:task_conflicts}. The Figure~\autoref{fig:tc:asr} demonstrates the gradient similarity in ASR pre-training.  
In most layers, the gradient similarities are small. No serious gradient interference is observed.
The Figure~\autoref{fig:tc:st} depicts the gradient similarity from the ST pre-training. Compared with the ASR pre-training, the S2T and T2T 
subtasks are replaced by speech translation and text based neural machine 
translation subtasks in pre-training.  The interference between different subtasks is significant as large positive and  negative gradient similarities are observed in the third and fifth layers in \autoref{fig:task_conflicts}. 

Similarly, we compare task gradients in the speech encoder and no obvious task interference is observed within the speech encoder for both ASR and ST pre-training. Detailed analysis on the speech encoder is included in the \autoref{sec:app:gsse}.

In order to alleviate the task interference,  the PSE configuration is proposed for the ST pre-training. \autoref{tab:asr_config_cmp} presents the performance comparison between two configurations on both ASR and ST pre-training.
On the left part of the table, we list the ASR results using 100 hours labelled speech data (train-clean-100) in pre-training and fine-tuning.
While the right part of the table shows the BLEU evaluated on the \textsc{MuST-C} dataset.
As we expected,  the FSE configuration encourages information sharing among tasks and it achieves lower WER for the ASR task. It indicates subtasks in the ASR pre-training are complementary to each other. On the other hand, the PSE configuration minimizes the information sharing between AED subtasks and encoder only subtasks, and it leads to higher BLEU for the ST task.

\begin{table}[]
    \centering
    \small
    \begin{tabular}{l|c|c||c|c}
    \toprule
    \multirow{2}{*}{Config.} & \multicolumn{2}{c||}{Librispeech (WER~$\downarrow$) } & \multicolumn{2}{c}{MuST-C (BLEU $\uparrow$)} \\
    \cline{2-5}
    & dev clean & dev other & EN-ES & EN-FR \\
    \hline
          FSE & 3.2 & 6.8 & 31.4 & 38.3\\ 
          PSE & 3.1 & 8.3 & 33.1 & 39.7 \\ 
    \bottomrule
    \end{tabular}
    \caption{ Comparison of two pre-training configurations for ASR and ST.}
    \label{tab:asr_config_cmp}
\end{table}

\subsection{Impact of training data}
The supervised speech data connects the text and speech modeling and unifies the representation from different modalities. 
An interesting question we want to investigate is how much supervised data is enough to learn a good cross modality representation. 
In this experiment, we choose different amounts of labelled data for ASR pre-training and fine-tuning, varied from 960 hours (the full dataset),  100 hours (train-clean-100)  and 10 hours as~\cite{librilight}, to answer this question. 

In \autoref{tab:cmp_data}, the first column shows the amounts of supervised speech data available during the pre-training  and the 
second column presents the amount of labelled data used in the fine-tuning stage. In pre-training, the same supervised speech data 
is used in the S2P and S2T subtasks. 

The first observation is that more supervised speech data in the pre-training stage is always helpful to get smaller WER.  
For example, if the models are fine-tuned with the full \textsc{Librispeech} training dataset, the average WER are 3.3 (row one), 3.6 (row two) and 4.0 (row four) for experiments with 960, 100 and 10 hours labelled data in the pre-training stage. 
The second observation is that we are still able to obtain good speech representations even with small amounts of labelled data. 
In row four, the model is pre-trained with 10 hours labelled data, then fine-tuned with 960 hours supervised speech data. 
It can achieve an average 4.0 WER, which is better than the results of the AED systems in \autoref{tab:asr}. However, we also notice the performance degrades quickly if only small amounts of labelled speech data are available. The average WER is increased to 24.6 (row six) when only 10 hours of supervised speech data is employed in both pre-training and fine-tuning.
  
\begin{table}[]
    \centering
    \small
    \begin{tabular}{c|c|c|c|c|c}
        \toprule
         \multirow{2}{*}{PT (h)} & \multirow{2}{*}{ FT (h)}   &\multicolumn{2}{c|}{Dev} & \multicolumn{2}{c}{Test} \\
         \cline{3-6}
         & & clean & other & clean & other  \\
         \hline
         \hline
         960 & 960 & 2.0 &  4.4 & 2.1 & 4.6 \\
         \hline
         \hline
         \multirow{2}{*}{100} & 960 & 2.3 & 4.9 & 2.2 & 5.1 \\
         \cline{2-6}
             & 100 & 3.2 & 6.8 & 3.5 & 7.2\\   
         \hline
         \hline
         \multirow{3}{*}{10} & 960 &2.7 & 5.3 & 2.8 & 5.3 \\
         \cline{2-6}
            & 100 & 3.8 & 7.8 & 4.0 & 7.7 \\ 
         \cline{2-6}
            & 10 &19.9 & 27.5 & 22.0 & 28.8 \\ 
        \bottomrule
    \end{tabular}
    \caption{Impact of the amounts of supervised data. ``PT'' and ``FT'' stand for pre-training and fine-tuning respectively.}
    \label{tab:cmp_data}
\end{table}

Another question we are interested is the generalizability of the pre-trained model. 
 There are two data partitions in \textsc{Librispeech}: ``clean'' and ``other''. The ``clean'' partition is supposed to be ``higher recording quality and with accents closer to US English'' while the ``other'' partition is difficult speakers with high WER~\cite{Panayotov2015LibrispeechAA}. 
We create four data partitions for pre-training and fine-tuning to simulate the mismatched training conditions. ``train-clean-100'' is used as the pre-training ``clean'' data set (``PT C'') and the first 30,000 utterance from ``train-clean-360'' as the fine-tuning ``clean'' dataset (``FT C''). The first 30,000 utterances and the following 30,000 utterances from ``train-other'' are used as the pre-training (``PT O'') and fine-tuning ``other'' (``FT O'') datasets. Each dataset includes approximately 100 hours speech data. 
In \autoref{tab:conditions}, models are trained under 4 different combinations with different supervised pre-training and fine-tuning data sets. We report average WER on the ''dev-clean'' and ``test-clean'' test sets as ``clean'',   and average WER on the``dev-other'' and ``test-other'' as ``other'' to reduce the result variation. 
From \autoref{tab:conditions}, we have following observations. 
1) a model achieves the best results on the matched condition. The model ``PT C + FT C'' achieves the lowest WER on the ``clean'' set while ``PT O + FT O'' achieves the best results on the ``other'' set. 
2) training and test on totally different conditions could increase WER significantly. The model ``PT C + FT C'' increases 0.9 WER on the ``other'' set compared with the ``PT O + FT O'' model. 3) mismatched pre-training and fine-tuning might slightly increase the WER, 0.1 to 0.2 in this experiment.

%

\begin{table}[]
    \centering
    \small
    \begin{tabular}{c|c|c||c|c}
    \toprule
      & \multicolumn{2}{c||}{FT C} &  \multicolumn{2}{c}{FT O}   \\
    \cline{2-5}
       &  clean &  other &  clean &  other \\
     \hline
     PT C & 3.0 & 6.7 & \textit{3.2} & \textit{5.9} \\ 
     \hline
      PT O & \textit{3.0} & \textit{5.9} & 3.2 & 5.8 \\ 
    \hline
    \bottomrule
    \end{tabular}
    \caption{WER comparison under mismatched pre-training and fine-tuning conditions. ``C''  and ``O'' represent the ``clean'' and ``other'' labelled data;  ``PT'' and ``FT'' stand for pre-training and fine-tuning.  WERs obtained under mismatched conditions are shown as \textit{italic}. }
    \label{tab:conditions}
\end{table}

\subsection{Masked KL divergence loss v.s. contrastive loss }
In the SSL subtask, we optimize the model to reduce the KL divergence loss between input without masking and with masking as described in \autoref{sec:method:ssl}. It is a variant of the masked prediction loss~\cite{Baevski2020vqwav2vecSL} and no target labels are required in our implementation. Contrastive loss is another widely used method for the self-supervised speech learning~\cite{Baevski2020wav2vecV2}. 
We compare the both criteria in \autoref{tab:loss}.  The number of distractors in the contrastive loss is 100 as  ~\cite{Baevski2020wav2vecV2}. 
Both ASR and ST results are reported in \autoref{tab:loss}, where the masked KL divergence loss achieves about $0.6$ lower WER in the Librispeech dev sets and $0.7\sim 1.4$ more BLEU scores in the MuST-C tst-COMMON sets. It demonstrates the effectiveness of the proposed masked KL divergence loss for the SSL subtask.
\begin{table}[]
    \centering
    \small
    \begin{tabular}{l|c|c||c|c}
    \toprule
    \multirow{2}{*}{Loss} & \multicolumn{2}{c||}{Librispeech (WER~$\downarrow$) } & \multicolumn{2}{c}{MuST-C (BLEU $\uparrow$)} \\
    \cline{2-5}
    & dev clean & dev other & EN-ES & EN-FR \\
    \hline
        Cont.  & 2.6  & 5.0  &  31.7 & 39.0 \\
        KL &  2.0   & 4.4   &  33.1 &  39.7 \\
    \bottomrule
    \end{tabular}
    \caption{Comparison of the masked KL divergence loss and contrastive loss for the SSL subtask. ``Cont.'' stands for the contrastive loss.}
    \label{tab:loss}
\end{table}

\subsection{Ablation study}
In \autoref{tab:ablation}, we present an ablation study  by removing different steps/tasks in the pre-training stage.  

In order to make the pre-training more stable, the model training adopts a three-stage optimization strategy: 1) pre-training the T2T subtask to have a good initialization on the phoneme embeddings 2) joint pre-training with four subtasks to leverage large amounts of unlabelled speech data and abundant text data  and 3) fine-tuning the model on the downstream task for best performance.
In the second row, we skip the T2T pre-training step and initialize the model randomly for 
the joint pre-training. 0.5 WER increase is observed in average on two
\textsc{Librispeech} dev sets.  It also has more impact on the EN-ES  translation direction  where  {1.2} BLEU score is lost without proper initialization. 

In the third row, we present the results without the S2T subtask. For both ASR and ST, significant performance degradation is observed,  with an average 1.1 WER increase for two ASR tests and 1.8 BLEU decrease for two ST directions.  
We also try removing the S2P subtask while still keeping the S2T subtask. The training doesn't converge. The SSL subtask is with very small or zero cost since all predictions collapse into one or two target phonemes. Also, little progress has been made for the S2T subtask even though it is co-trained with the SSL and T2T subtasks. 

In the last row, the model is trained without pre-training, i.e., only the T2T and S2T subtasks are optimized. 
Compared with the \textsc{STPT} results,  there is about 1.4 WER increase for two \textsc{Librispeech} test sets  and 3.4 BLEU decrease for the two ST directions on average.

\begin{table}[]
    \centering
    \small
    \begin{tabular}{l|c|c||c|c}
    \toprule
    \multirow{2}{*}{Config.} & \multicolumn{2}{c||}{Librispeech~(WER $\downarrow$) } & \multicolumn{2}{c}{MuST-C~(BLEU $\uparrow$)} \\
    \cline{2-5}
    & dev clean &  dev other & EN-ES & EN-FR \\
    \hline
          STPT & 2.0 & 4.4 & 33.1 & 39.7\\  
          -  T2T PT & 2.4 & 5.0 & 31.9 & 39.2   \\ 
          - AED task & 2.9 & 5.6 & 31.3  & 38.0   \\  
          - Joint PT &2.8 & 6.4 & 30.6 & 35.4 \\
    \bottomrule
    \end{tabular}
    \caption{ Ablation study for STPT.``PT'' stands for ``pre-training''.}
    \label{tab:ablation}
\end{table}


\section{Conclusion}
In this work, we present a method to jointly pre-train speech and text in one model for speech translation and recognition under the AED framework. It includes four self-supervised and supervised subtasks from two different input modalities, hence the proposed method can leverage large amounts of unlabelled speech data and abundant text data in the pre-training stage.  
We conduct detailed analysis on the interference among different subtasks and propose two model configurations for the ASR and ST pre-training respectively to alleviate the subtask interference.
Our experimental results show STPT can effectively fuse information within text and speech training data into one model. We achieves between $1.7$ and $2.3$ BLEU improvement over the state of the art on the \textsc{MuST-C} EN-FR and EN-ES speech translation tasks, and comparable WERs as wav2vec 2.0 in the \textsc{Librispeech} ASR task.

\section{Acknowledgments}
We want to thank the anonymous reviewers for their insightful comments and suggestions.

\section{Broader Impact} We highlight the potential that this work has positive impact in the society: augmenting speech processing tasks with text corpus, and improving speech related applications. At the same time, this work may have some negative consequences if the text data is not handled in a proper way. Before using the text data to train a speech system, one should evaluate fairness in the collected data, and make sure not to train on offensive or any type of inappropriate data.
\newpage

\bibliographystyle{acl_natbib}
\bibliography{anthology,acl}

\newpage

\appendix

\begin{figure*}[t]
\hfill
\centering
\subfigure[Gradient similarity for the ASR pre-training.]{\includegraphics[width=1.0\columnwidth]{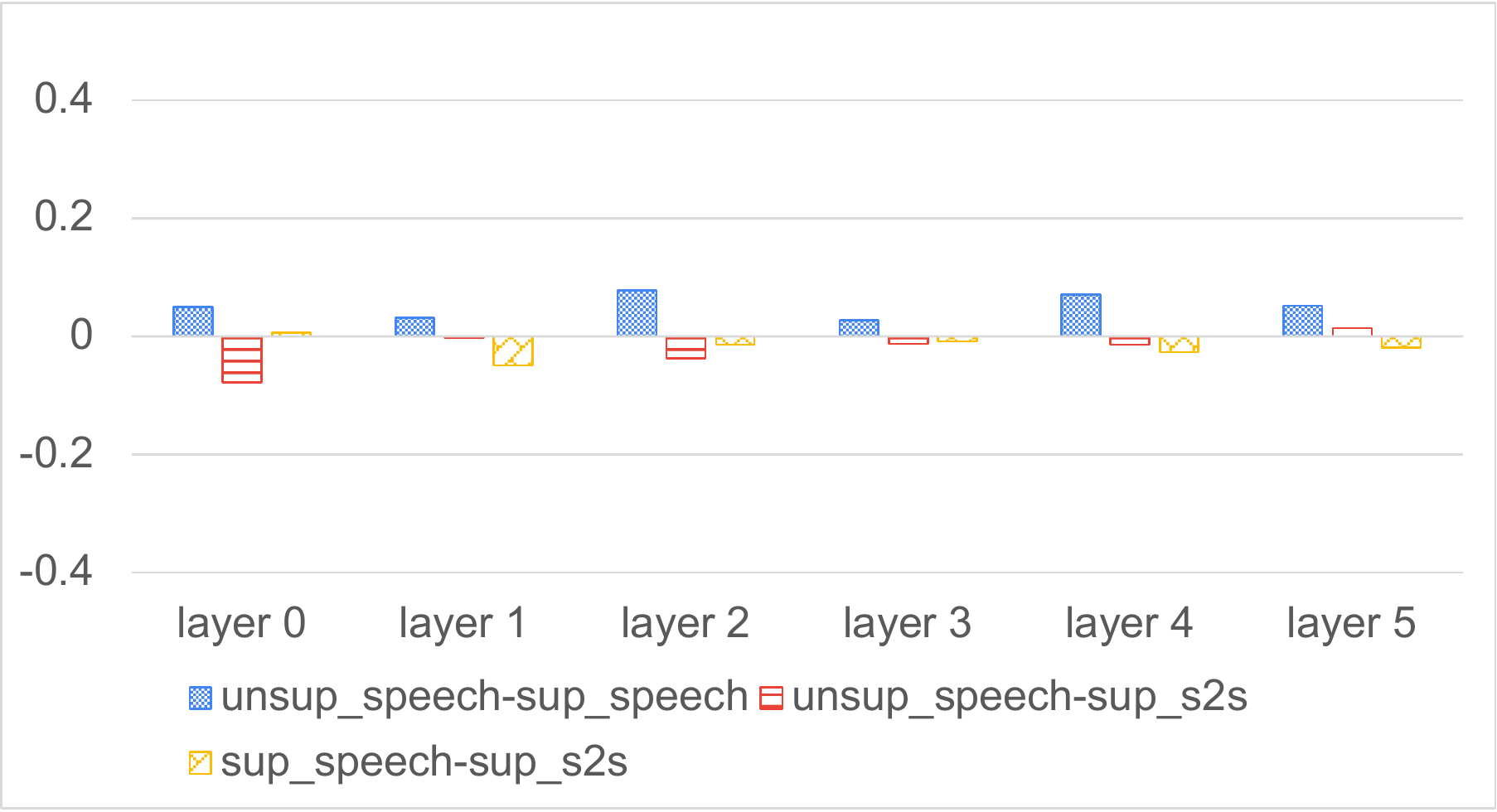}}
\hfill
\subfigure[Gradient similarity for the ST pre-training.]{\includegraphics[width=1.0\columnwidth]{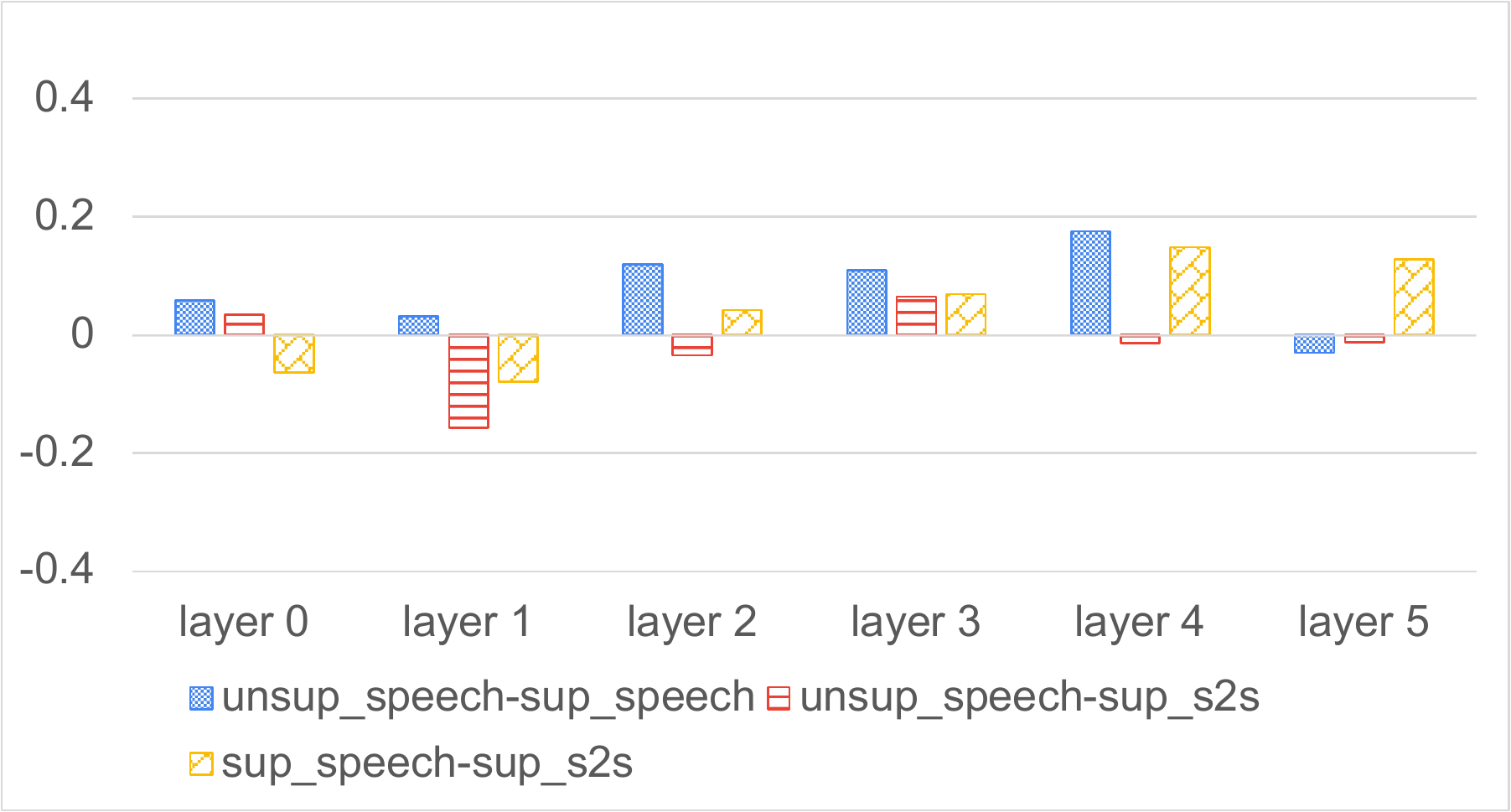}}
\hfill
\caption{Gradient similarity for different subtasks on the speech encoder.}
\label{fig:speech_task_conflicts}
\end{figure*}

\section{Pre-training data setting}\label{sec:data}
\noindent\textbf{T2T}: For ASR pre-training, the language model~(LM) training dataset~\footnote{https://www.openslr.org/11/} for \textsc{Librispeech}~\cite{Panayotov2015LibrispeechAA} is used to build the monolingual BART model.  It has about 800 million words. For ST pre-training, we take the parallel training corpus from WMT. 
We examine our methods on two translation directions in \textsc{MuST-C}: 
English-Spanish (EN-ES), which uses WMT13 training corpus,  and English-French (EN-FR), which takes the WMT14 training data. 
There are 370 million and 1 billion English words in the EN-ES and EN-FR parallel training datasets respectively. 

We use ``g2p\_en'' Python package~\cite{Lee2018LearningPF} to convert the training text into the corresponding phoneme representation, which is based on the CMU English dictionary.  
We further extend the phoneme set by distinguishing the first phoneme in the word with an additional “\_” mark appended, which is similar to the notation in the SentencePiece process. The input phoneme vocabulary size is 134. 

\noindent\textbf{SSL}:
For both ASR and ST pre-training, 60k hours of unlabelled English speech data from Libri-light~\cite{librilight} is used to build the self-supervised speech task if not specifically mentioned. We set the maximum utterance duration to 37.5 seconds and minimum duration to 4 seconds. We  randomly sample  audio segments with maximum duration if utterances are longer than the maximum duration. No voice activity detection is applied.  

\noindent\textbf{S2P}:
We use the transcribed \textsc{Librispeech} dataset for ASR pre-training.  In ST pre-training, the \textsc{MuST-C} training dataset is used, where the corresponding English transcription is used as the training target labels after it is converted into phoneme representation. The phoneme level segmentation is obtained via force-alignment, which is conducted using HMM/GMM trained from the same speech data with the Kaldi toolkit~\cite{Povey_ASRU2011}.

\noindent\textbf{S2T}:
We use the same labelled data in the S2P subtask for the S2T subtask, i.e., \textsc{Librispeech} training data for the ASR pre-training and \textsc{MuST-C} data for the ST pre-training. 
 Instead of using phoneme representation, the target labels are encoded with SentencePiece~\cite{Kudo2018SentencePieceAS}. For both ASR and ST tasks, the vocabulary is an Unigram model with size 10k  and full character coverage on the training text data.  
 
\section{Optimization setting}\label{sec:app:opt}
The models are optimized with Adam~\cite{kingma2014adam} for both pre-training and fine-tuning. The final results are evaluated using an averaged model from checkpoints of the last 10 epochs. 
\textbf{T2T subtask pre-training} 
The T2T model is pre-trained with learning rate 0.01 using Adam optimization. The maximum tokens per mini-batch is 2048 with 8 V100 GPU cards.
The model is updated 400,000 until fully converged. 
\\
\textbf{Pre-training with all subtasks}
 The model then keeps optimizing with all four subtasks: T2T, SSL, S2P and S2T, with learning rate 0.001. The model is trained using 16 A100 GPU cards with update frequency 12. The maximum token number per batch for the T2T subtask is 2048 while the maximum sample number is 750,000 (46s) for the speech input in three speech subtasks. The  
 maximum update number is 800,000 and 200,000 for the ASR pre-training and the ST pre-training respectively.
\\
\textbf{Fine-tuning} The model is fine-tuned on the downstream task with learning rate 0.0003 and 8 V100 GPU cards. The update frequency set to 3.
The maximum update numbers are dependent on the amounts of  supervised speech data available. We choose 100,000 for the ASR task with 960 hours training data and 20,000 for 100 or 10 hours training data. For the ST task, the maximum update number is set to 50,000.

\section{Gradient similarity of the speech encoder}\label{sec:app:gsse}
Three subtasks: SSL, S2P, and S2T, share the speech encoder during the joint pre-training.
Similar pairwise gradient similarity analysis is conducted  on these three subtasks at the speech encoder, as shown in \autoref{fig:speech_task_conflicts}. The gradient similarity analysis for the ASR pre-training is presented in the left subfigure while the ST-pretraining is listed in the right. In both cases, the gradient similarities for different subtask pairs are small, i.e., absolute values of the gradient similarities are all below 0.2. It indicates the task interference between different subtasks are not significant.

\end{document}